\documentclass[10pt,twocolumn,letterpaper]{article}

\usepackage{iccv}
\usepackage{times}
\usepackage{epsfig}
\usepackage{graphicx}
\usepackage{amsmath}
\usepackage{amssymb}

\usepackage{multirow}
\usepackage{hhline}
\usepackage{textcomp}
\usepackage{caption} 
\usepackage{textcomp}
\usepackage{siunitx}
\usepackage{subcaption}
\usepackage{authblk}
\usepackage[utf8]{inputenc}
\usepackage{dirtytalk}

\usepackage[breaklinks=true,bookmarks=false]{hyperref}

\iccvfinalcopy 

\def\iccvPaperID{****} 

\begin{document}

\makeatletter
\newcommand{\printfnsymbol}[1]{%
  \textsuperscript{\@fnsymbol{#1}}%
}

\makeatother

\newcommand{\cref}[1]{\ref{#1}}
\newcommand{\pn}{Plugin Network }
\newcommand{\squeezeup}{\vspace{0.5em}}
\title{Plugin Networks for Inference under Partial Evidence}

\author[1]{Michal Koperski\thanks{Equal contribution}}
\author[2,3,1]{Tomasz Konopczyński\printfnsymbol{1}}
\author[4,1]{Rafa\l{} Nowak}
\author[5,1]{\\Piotr Semberecki}
\author[6,1]{Tomasz Trzciński} 

\affil[ ]{{\tt\small \{firstname.lastname\}@tooploox.com} \footnotesize}
\affil[1]{\footnotesize Tooploox Ltd., Poland}
\affil[2]{\footnotesize Department of Radiation Oncology, University Medical Center Mannheim, Heidelberg University, Germany}
\affil[3]{\footnotesize Interdisciplinary Center for Scientific Computing (IWR), Heidelberg University, Germany}
\affil[4]{\footnotesize Institute of Computer Science, University of Wroclaw, Poland}
\affil[5]{\footnotesize Wroclaw University of Science and Technology, Poland}
\affil[6]{\footnotesize Warsaw University of Technology, Poland}

\maketitle

\begin{abstract}
In this paper we propose a~novel method to incorporate partial evidence in the inference of deep convolutional neural networks.
Contrary to the existing, top performing methods, which either iteratively modify the input of the network or exploit external label taxonomy to take the partial evidence into account, we add separate network modules (\say{Plugin Networks}) to the intermediate layers of a~pre-trained convolutional network.
The goal of these modules is to incorporate additional signal, \ie~information about known labels, into the inference procedure and adjust the predicted output accordingly.
Since the attached plugins have a~simple structure, consisting of only fully connected layers, we drastically reduced the computational cost of training and inference.
At the same time, the proposed architecture allows to propagate information about known labels directly to the intermediate layers to improve the final representation.
Extensive evaluation of the proposed method confirms that our Plugin Networks outperform the state-of-the-art in a~variety of tasks, including scene categorization, multi-label image annotation and  semantic segmentation.
\end{abstract}

\section{Introduction}

Visual recognition tasks, \eg scene categorization or multi-label image annotation, have attracted a~significant amount of research interest in recent years~\cite{Wang18,Hu2016,Xiao10,Fang15}. One of the reasons, which sparked this attention was the availability of evaluation datasets created for benchmarking given visual tasks, such as ImageNet~\cite{imagenet}, VOC Pascal~\cite{pascal-voc-2011} or COCO~\cite{Lin14}. Although sensible for comparison purposes, single-task evaluation protocols are often far from real-life use-cases, where additional information, \eg related to location or time of photo capture, is available.  
\begin{figure}[t]
\begin{center}
    \includegraphics[width=0.99\linewidth]{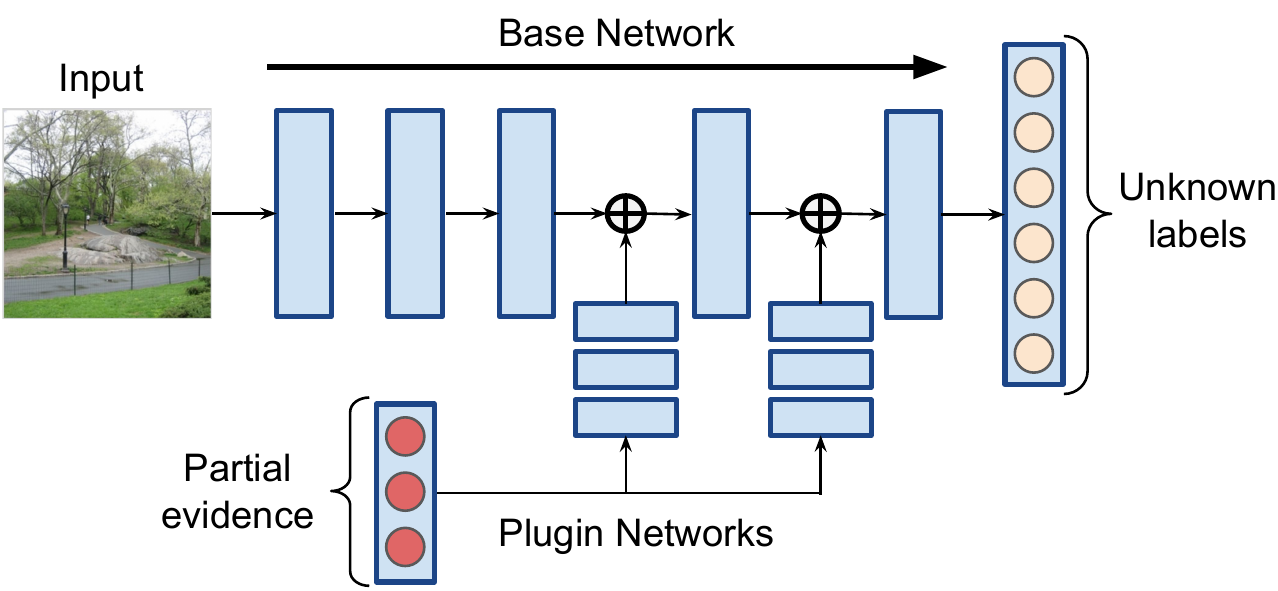}
\end{center}
\caption{Plugin Networks --- neural networks attached to the intermediate layers of a~pre-trained convolutional neural network, allow to exploit partial evidence labels at the inference to predict unknown labels with higher accuracy. This simple, yet effective approach significantly reduces train and test time, while outperforming competitive results on three challenging benchmarks.
}
\squeezeup
\label{fig:overview}
\end{figure}

The availability of partial information ({\it partial evidence}) about an image, made available at test time, can improve accuracy of pre-trained networks~\cite{Hu2016,Wang18}, and we will follow this scenario. More specifically, we assume that a~set of labels corresponding to a~given image is known during inference, while the task at hand is to improve the performance of the model on the original task, \eg image classification, object detection, semantic segmentation. This corresponds to a~real life application, where, for instance, we know that the image was captured in a~forest or in a~cave, which drastically reduces the likelihood of detecting a~skyscraper. Similarly, information that a~given object appears in an image can greatly improve its localization or segmentation. Since partial evidence can be available in multiple forms and modalities, the main prediction system, \eg a~convolutional neural network (CNN), is trained to perform a~general purpose prediction with no assumption about the existence of partial evidence or lack thereof. Neural architectures such as CNNs are not modular, thus any modification such as new inputs (partial evidence) or new outputs (new tasks) are difficult to apply without repeating the full training procedure. Otherwise, phenomena such as the catastrophic forgetting may occur. Our objective is to enable the model to incorporate additional available information without re-training the main system while exploiting this information to increase the quality of predictions.

Several methods were proposed in the literature to address this problem, among them~\cite{Hu2016} and~\cite{Wang18} are the most recent. In paper \cite{Hu2016} authors proposed to exploit external taxonomy of the labels by modelling correspondences between scene attributes and categories, by feeding this data into the main neural network at inference. On the other hand, paper \cite{Wang18} introduces a~feedback-prop approach that iteratively modifies input and network activations to ensure the response of the network corresponds to the distribution of known labels. Although, those methods provide effective ways to exploit partial evidence, they require complex reasoning, that concerns relationships between labels or computationally expensive iterative adaptation mechanism.

In this work, we reduce the complexity and propose the~Plugin Networks -- a~simple, yet effective main network extension that allows to incorporate partial evidence during the inference. We show that by using a~set of fully-connected (FC) side networks attached to intermediate layers of the main network (see Fig.~\ref{fig:overview}), we are able to not only avoid costly optimization process, but also exploit the assumption about the existence of partial evidence in the offline training stage. More specifically, the proposed Plugin Networks, connected to the backbone neural network, adjust their activations at the time of inference, depending on available known labels. Due to the simplicity of the Plugin Networks, their training converges quickly, while remaining robust to overfitting, as we show in this paper. The inference of the proposed model consists of a~quick feed-forward propagation of the main model. Plugin Networks offer a~significant speedup with respect to the state-of-the-art feedback-prop method~\cite{Wang18}. Last but not least, the proposed Plugin Networks outperform all of the existing methods on three challenging benchmark applications: hierarchical scene categorization on the SUN397 dataset~\cite{Xiao10}, multi-label image annotation on the COCO 2014 dataset~\cite{Lin14}, and semantic segmentation on Pascal VOC 2011~\cite{pascal-voc-2011}.

To summarize, our contributions are as follows:
\begin{itemize}
    \item We propose novel neural network model extensions called Plugin Networks, which allows us to take partial information available at test time into account. Plugin Networks adjust the activations of the pre-trained base network. They are fast to train and efficient at test time.
    \item We show how to attach the proposed Plugin Networks to different types of neural network layers and investigate the influence of those variants on final results.
    \item We provide an extensive evaluation of the proposed approach on three challenging tasks: hierarchical scene categorization, multi-label image annotation and scene segmentation.
    
\end{itemize}
We make our code available for the public\footnote{\url{https://github.com/tooploox/plugin-networks}}.

In the remainder of this paper, we first give an overview of related publications. In Sec.~\ref{sec:method} we formally introduce the proposed approach, explain how to use it and discuss its properties. Sec.~\ref{sec:experiments} provides an extensive evaluation of our method and we conclude this paper in Sec.~\ref{sec:conclusions}.

\section{Related Work}
\label{sec:related}
\noindent\textbf{Using context in visual tasks}: Exploiting additional contextual cues in visual recognition tasks gained a~lot of attention from the computer vision community~\cite{Divvala09,Galleguillos08,Johnson15}. Contextual information related to semantics was used to improve object detection~\cite{Mottaghi14}. Social media meta-data was also used in a~context of multilabel image annotation in~\cite{Johnson15}. Although, adding context proved to be successful in increasing the quality of visual recognition tasks, all of the above mentioned methods used the context in conjunction with the input uni-modal (visual) image during the training of the entire system. In this work, we propose a~fundamentally different approach since the context (in the form of known labels) is learned only after the training of the main model is finished and our approach allows to extend this pre-trained model with additional information {\it a~posteriori}. Rosenfeld \etal~\cite{Rosenfeld18} proposed a~method where detection and segmentation is conditioned on the presence of a~given object category. To achieve it, they propose to use a~set of linear modulators. This method shares common features such as offline training with the Plugin Networks, but a~model capacity of linear modulators is not enough to learn complex functions. This leads to more complicated training procedures with data oversampling.
Perez \etal~\cite{film} proposed FiLM (Feature-wise Linear Modulation) for visual question answering, where textual information serves as a context to a CNN model.
In particular FiLM introduces new layers (named FiLM blocks or Resblocks), which are incorporated into the base model.
FiLM blocks are later modulated using affine fusion operator.
Plugin Networks on the other hand do not introduce any alternations to the base model architecture and directly modulate activations of the base model.
Thanks to that, we do not introduce so much noise as the FiLM Resblocks to the base model which results in a more stable training.

\noindent\textbf{Using label structure}: Some authors proposed to model the co-occurrence of labels available at training time to improve recognition performance~\cite{Miech17}. \cite{Deng14} on the other hand uses a~special structure to store the relations between the labels using a~graph designed specifically to capture semantic similarities between the labels. Other forms of external knowledge can be found in~\cite{Grauman11} and~\cite{Hwang12} where they use the~WordNet taxonomy of tags to increase the accuracy of their visual recognition systems. \cite{Johnson15}, \cite{McAuley12} also used social media meta-data to improve the quality of the results obtained for image recognition tasks. Finally, ~\cite{Ordonez13} estimated entry-level labels of visual objects by exploiting image captions. Contrary to our method, the above-mentioned approaches focus on finding the relationships between the labels and driving the training algorithm to encompass those structures. In this work, we do not explicitly model any label structures -- the only input related to labels we give to the network is a~set of known labels related to an image with no information about their relationship with the others. 

\noindent\textbf{Multi-task learning}: Somehow related to our work is the thriving area of multi-task learning. Motivated by the phenomenon of catastrophic forgetting, multi-task learning tries to address the problem of lifelong learning and adaptation of a~neural network to a~set of changing tasks while preserving the network's structure. In \cite{Lee17}, Lee \etal aim to solve this problem by the continuous matching of network distribution. In~\cite{Rebuffi17} the same problem is solved through residual adapters -- neural network modules plugged into a~network, similarly to our Plugin Networks -- which are the only structures trained for the tasks while the base network remains untouched. Although, we do not aim to solve multi-task learning problem in this work, our approach is inspired by the above-mentioned methods, which focus on designing robust network architecture that can dynamically adjust to additional data point sources unseen during training.

\noindent\textbf{Inference with Partial Evidence:}
Finally, the most relevant to the work presented in this paper are two methods proposed by Hu \etal \cite{Hu2016} and Wang \etal \cite{Wang18}. Both of them address the problem of visual tasks in the presence of partial evidence. 

Hu \etal~\cite{Hu2016} tackles this challenge by proposing a~Structured Inference Neural Network (SINN). The SINN method is designed to discover the hierarchical structure of labels, but it can also be used in a~partial evidence setup, if labels in a~given hierarchy are clamped at inference. However, the SINN model, which uses CNN and LSTM to discover label relations, has a~large amount of learnable parameters, which makes model training difficult. To solve this issue authors use the positive and negative correlations of labels as prior knowledge, which is inferred from the WordNet relations. We compare our method with SINN and show that we achieve significantly better performance with a~much simpler model.

The FeedbackProp proposed by Wang \etal \cite{Wang18}, uses an iterative procedure, which is applied at inference time. The idea is to modify network activations to maximize the probabilities of labels under the partial evidence. The method does not require to re-train the base model. However, due to the iterative procedure introduced at inference time, it requires more computational effort. In addition, they introduced hyperparameters, like a~number of iterations and learning rate to the inference phase. Finally, in the case of FeedbackProp, the partial evidence labels can only be a~subset of labels that the base model can recognize. Our method, however, can accept any kind of labels as partial evidence. Moreover, our method introduce negligible computations, and not extra parameters to the inference phase.
The comparison shows that our method outperforms FeedbackProp while being significantly faster at inference phase.
\section{Plugin Networks}
\label{sec:method}
In this section, we first introduce the Plugin Networks and define them formally. We then describe how to attach the Plugin Networks to the existing base network at the linear and convolutional layers.

\subsection{Definition}

Let's assume that we have a~CNN model $F(\mathbf{x}; \mathbf{w})$, where $\mathbf{x}$ is an input image and $\mathbf{w}$ are the parameters. The model $F$ is already trained on some task (\eg single or multi-label classification, scene segmentation). The parameters $\mathbf{w}$ were trained on input images $X$ and input labels $Y$.

Now let's assume that some labels $\bar Y$ are available and known
at inference time. 
In the following definitions without loosing generality, we will assume that only one Plugin Network is attached to the base model. 
We define the Plugin Network model $F_p$ with parameters $\mathbf{w}_p$ as
\begin{align}
    \mathbf{r} = F_p(\mathbf{\bar y};\mathbf{w}_p).
\end{align}
The model takes the partial evidence $\mathbf{\bar y} \in \bar Y$ as an input. The output $\mathbf r$ of the plugin can be attached to the output vector $\mathbf{z}$ of some layer of the base model $F$:
\begin{align}
    \mathbf{\tilde{z}} = \mathbf{z} \oplus \mathbf{r},
    \label{eq:plugin}
\end{align}
where the sign $\oplus$ can have the following meaning:
\begin{itemize}
    \item additive: $\mathbf{\tilde{z}} = \mathbf{z} + \mathbf{r}$,
    \item affine: $\mathbf{\tilde{z}} = \mathbf{r_a}\mathbf{z} + \mathbf{r_b}, \mathbf{r} = \mathbf{r_a} \mathbin\Vert \mathbf{r_b}$, 
    \item multiplicative: $\mathbf{\tilde{z}} = \mathbf{z} * \mathbf{r}$,
    \item residual:  $\mathbf{\tilde{z}} = \mathbf{z} + \mathbf{z}* \mathbf{r}$,
\end{itemize}
where $\mathbin\Vert$ is the concatenation operator.

In this way the Plugin Network $F_p$ adapts the output vector $\mathbf z$ of the base model $F$ under presence of available partial evidence.
The eq.~\eqref{eq:plugin} defines how the Plugin Network $F_p$ is attached  to the base network $F$, thus a~joint model can be defined as:
\begin{align}
    \tilde{F}(\mathbf{x},\mathbf{\bar y}; \mathbf{w},\mathbf{w}_{p_{i}}).\label{eq:join-model}
\end{align}
In general, several Plugin Networks can be attached simultaneously to  a~number of layers of the base model $F$.

Note that the output of a Plugin Network $\mathbf r$ can be attached to either the output of a~fully connected layer or a~convolutional layer.
In the following sections, we explain how both operations are performed in details.

\subsection{Connection with Linear Layers}
The Plugin Network, which is attached to the linear layer, has to compute the vector $\mathbf{r}$ of the same dimension as the vector $\mathbf{z}$. Then the operator $\oplus$ in~eq.~\eqref{eq:plugin} is well defined.
The adjusted vector $\mathbf{\tilde z}$ is then processed by the following layers of the base model.
The value $\mathbf{z}$ is the output of a layer before a~non-linear function (\eg ReLU).

\subsection{Connection with Convolutional Layers}
When a Plugin Network is attached to a~convolutional layer, it adjusts the feature map obtained from a~given convolutional filter. Thus, it has to compute vector $\mathbf{r}$, which has to be of the same dimension as the number of channels in the tensor $\mathbf{z}$.
Then all the considered operators $\oplus$ in eq.~\eqref{eq:plugin} have elementwise meaning:
\begin{align}
    \mathbf{\tilde z}_c &= \mathbf{z}_c \oplus r_c = \begin{bmatrix} 
    z_{11} &  \dots &z_{1j}\\
    \vdots & \ddots & \\
    z_{i1}  &  &  z_{ij} 
    \end{bmatrix}_c \oplus r_c,
\end{align}
where $c$ indicates the number of channel.

\begin{figure}[t]
\begin{center}
    \includegraphics[width=0.99\linewidth]{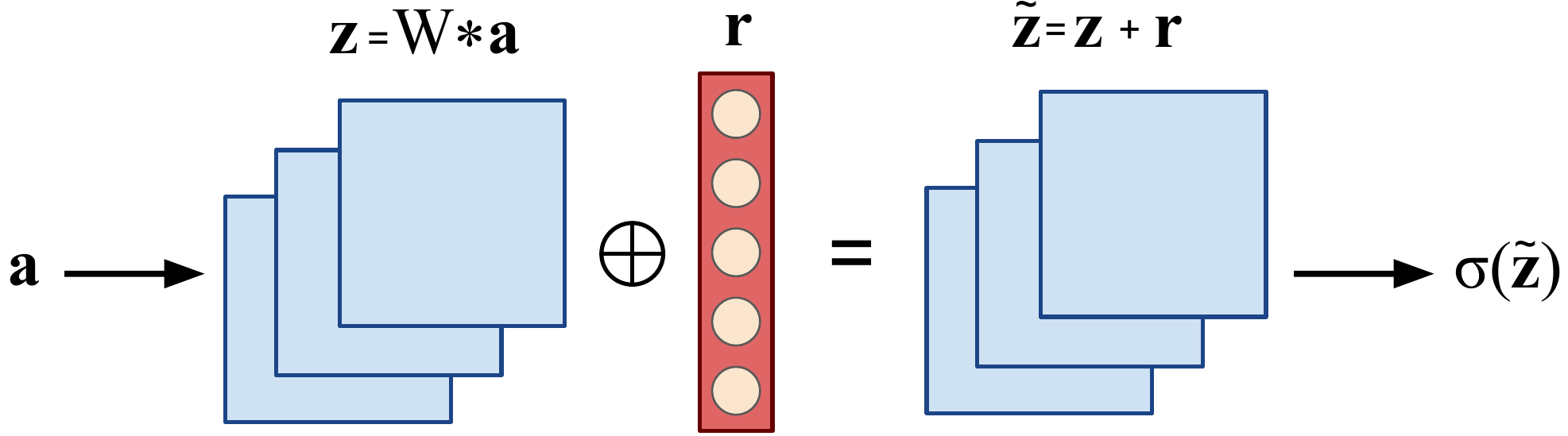}
\end{center}
\caption{If we add Plugin Network to a convolutional layer, then each element of the output vector is added to a~corresponding channel of feature maps. For instance, if convolutional layer has $c$ channels, then output from the Plugin Network has also $c$ elements.}
\squeezeup
\label{fig:method-1}
\end{figure}

Adding a~scalar value to each channel of a~feature map greatly reduces the~number of Plugin Network parameters that have to be learned.
Learning different values for each elements of a~feature map requires $w \times h \times c$ output values, where $w$, $h$ stands for width and height of a~feature map, respectively.
Since we only add scalar value to each feature map, we require only $c$ output values from a Plugin Network. Fig.~\ref{fig:method-1} illustrates how Plugin Network is attached to convolutional layers. 

\subsection{Plugin Network Architecture}

Overall, the Plugin Networks can be generalized to any model that can be trained with backpropagation. In our case, the Plugin Network is a~FC~neural network. Each FC layer is followed by a ReLU activation except for the last layer. We chose a~fully connected architecture because partial evidence vector $\mathbf{\bar y}$ can be interpreted as a~feature vector, which is used to compute a~non-linear transform of outputs of the base model. This task can be well handled by a~fully connected neural network.

\subsection{Training}

\label{sec:plugin:training}
The eq.~\eqref{eq:join-model} defines joint model of a base network $F$ with parameters $\mathbf{w}$ and the Plugin Network $F_p$ with parameters $\mathbf{w}_p$.
In the training procedure we are optimizing only $\mathbf{w}_p$ parameters, thus the base model $F$ is not altered.
To optimize the parameters of the Plugin Networks, the original loss function is used, \ie~the same loss function that was used to train the base model.

\subsection{Properties}
One important property of the Plugin Networks is that the function, which modifies a base model, is trained in an offline phase (see Section~\cref{sec:plugin:training}).
Thus, the testing phase requires only a~single feed-forward propagation through base model and the Plugin Networks.
Thanks to the single forward pass, our model is fast, and forward propagation overhead of the Plugin Networks is negligible.
Thus, our method is significantly faster than the model proposed in \cite{Wang18}, where iterative optimization process is applied at the inference phase. 

\section{Experiments} 
\label{sec:experiments}

In this section we evaluate the Plugin Networks on three challenging computer vision tasks.
We consider a~hierarchical, multi-label classification and semantic segmentation problems.
To stay consistent with the previous work on these subjects, we conduct the experiments in the same setups as \cite{Hu2016, Rosenfeld18, Wang18}.
Therefore, we use SUN397~\cite{Xiao10, Xiao14}, COCO'14 \cite{Lin14} and Pascal VOC 2011 \cite{pascal-voc-2011} datasets.

\subsection{Hierarchical Scene Categorization}
\label{sec:sun397}
We apply our method on SUN397 dataset \cite{Xiao10, Xiao14}.
The dataset is annotated with 3 coarse categories, 16 general scene categories and 397 fine-grained scene categories.
Our task is to classify fine-grained categories, given true values for coarse categories, as it was performed in Hu \etal \cite{Hu2016} and Wang \etal \cite{Wang18}. Thus, coarse categories serve as the partial evidence.
We follow same experimental setup as~\cite{Hu2016, Wang18}: we split dataset into train, validation, and test split with 50, 10 and 40 images per scene category.
To allow fair comparison to \cite{Hu2016, Wang18}, we use the AlexNet~\cite{Krizhevsky12} with Softmax trained on fine-grained categories.
It will serve as the base model for the Plugin Networks in this experiment.
To evaluate our method we compute mean average precision (mAP), multi-class accuracy (MC Acc) and intersection-over-union accuracy (IoU Acc).

\begin{figure}[t]
\begin{center}
    \includegraphics[width=0.81\linewidth]{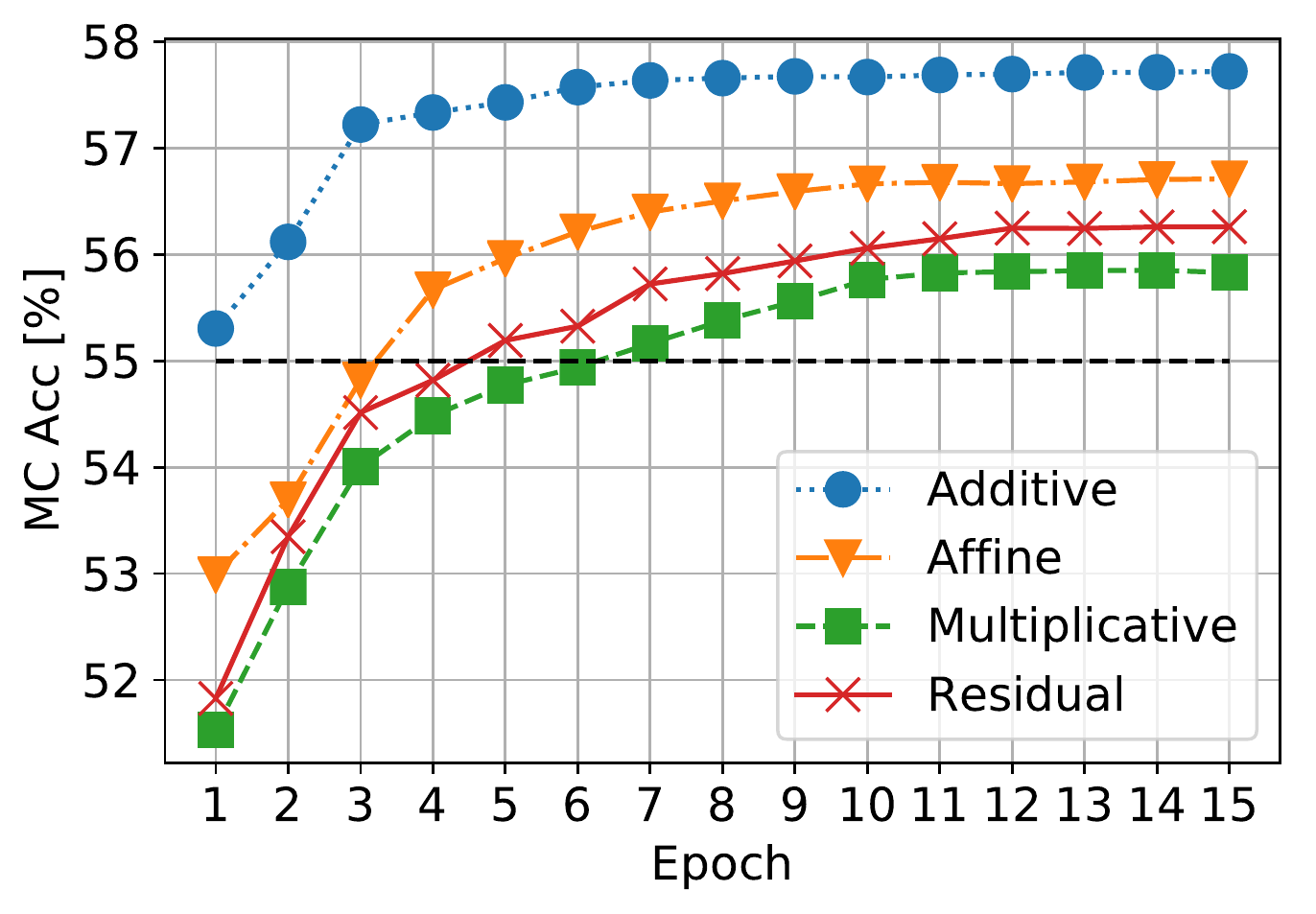}
\end{center}
\vspace{-2em}
\caption{
Plugin Networks -- fusion operators ablation study on SUN397 dataset.
Black dashed line marks state-of-the-art.}
\squeezeup
\label{img:ablation:FusionOperator:sun397}
\end{figure}

\begin{table}[t!]
\centering
\begin{tabular}{|c|c|}
\hline
Layer   & {MC Acc}\\ \hline
no plugin    & 53.15              \\ \hline
conv1   & 54.08      \\ \hline
conv2   & 53.88              \\ \hline
conv3   & 53.75                             \\ \hline
conv4   & 54.30                \\ \hline
conv5   & 54.86       \\ \hline
conv3-5 & 56.88      \\\hline
\end{tabular}
\quad
\begin{tabular}{|c|c|}
\hline
Layer   & {MC Acc}\\ \hline
{fc1}            & 53.90         \\ \hline
fc2              & 56.47           \\ \hline
fc3            &{\textbf{57.51}}  \\ \hline
fc1-3          & {57.08}                \\ \hline
conv3-5, fc1-3 & 57.16              \\ \hline
\end{tabular}
\caption{Performance of the Plugin networks \wrt the number of plugins and layers at which they are attached to the AlexNet. The comparison is done on SUN397 dataset.
The performance is better if plugin is attached to deeper layers. Plugins attached to FC layers perform better. The most effective is plugin attached to fc3 layer, although it outperforms slightly models where 3 and 6 plugins were attached. We believe that it is characteristic for classification task.}
\squeezeup
\label{tab:ablation:layer:sun397}
\end{table}

\begin{table}[t!]
\centering
\begin{tabular}{|l|l|r|r|r|r|}
\hline
\multirow{2}{*}{Layer} & \multicolumn{5}{c|}{\# of hidden layers}                                                                                           \\ \cline{2-6} 
                       & \multicolumn{1}{c|}{0}                          & \multicolumn{1}{c|}{1} & \multicolumn{1}{c|}{2} & \multicolumn{1}{c|}{3} & \multicolumn{1}{c|}{4} \\ \hline
conv5                  & \multicolumn{1}{r|}{53.41} & 54.06                  & 54.18                  & \textbf{54.28}                  & 53.90                  \\ \hline
fc3                    & \multicolumn{1}{r|}{54.53} & 56.65                  & 57.18                  & \textbf{57.51}                  & 56.56                  \\ \hline
\end{tabular}
\caption{
Performance of the Plugin Network (MC Acc) with respect to different number of hidden layers on SUN397.
We consider two cases. Former is the Plugin Network being attached to the conv5 layer, while the latter --- attached to the fc3 layer. The experiment shows that simple linear transformation (0 hidden layers) is not sufficient. 
}
\squeezeup
\label{tab:ablation:depth:sun397}

\end{table}

\noindent\textbf{Ablation study:} Plugin Network can be attached to different layers of the base model.
Furthermore, one can attach more than one Plugin Network simultaneously.
Now, we analyze different combinations of the above choices, the results are summarized in Table~\ref{tab:ablation:layer:sun397}.
The results show, that Plugin Network improve performance of the base model regardless to which layer it is attached.
On the other hand, the performance gain differs between chosen layers.
Thus, couple of observations can be drawn.
It is more effective to attach a Plugin Network to an FC layer rather than a conv layer.
Secondly, the Plugin Network connected to deeper layers tends to obtain better performance.
The results are aligned with the intuition, that in case of the classification task, deeper layers carry more abstract information.
Thus, the Plugin Network can converge to better solution when attached to deeper layers.
Moreover, in classification task spatial information is ignored, thus modification of conv layers, which carry spatial information is less important. 
Note, that in case of semantic segmentation task described in Sec.~\ref{sec:semantic:segmentation}, it is more important to modify conv layers. These experiments show that the Plugin Networks are generic and can solve various tasks.
We do not observe further performance improvements, if more than one Plugin Network is attached simultaneously.
In case of classification task, the Plugin Network mainly learn relationship between partial evidence and output labels.
Thus, fc3 outputs carry enough information to find such a~relationship. 

\begin{figure*}[t]
\begin{center}
    \includegraphics[width=0.9\linewidth]{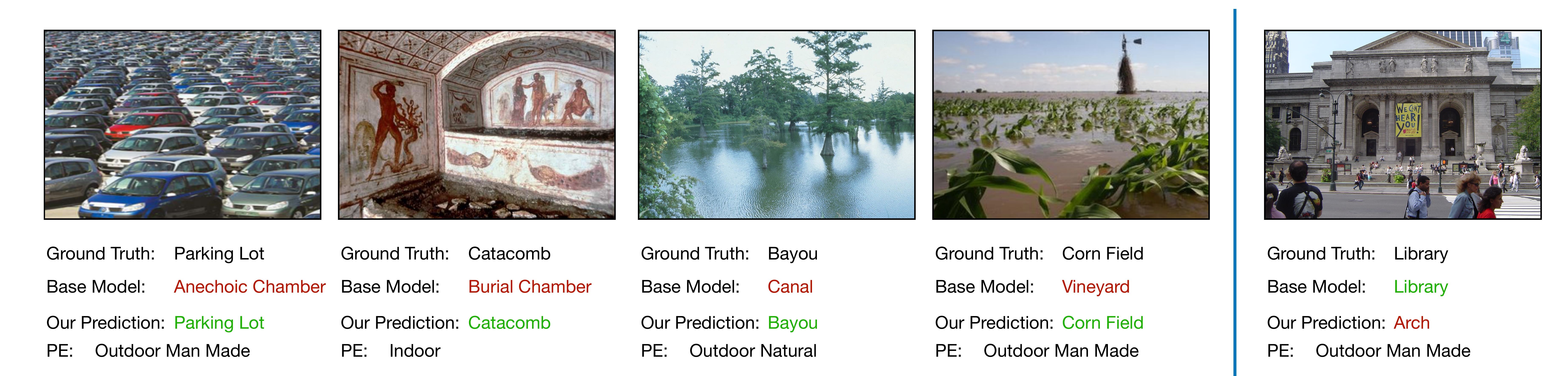}
\end{center}
\squeezeup
\caption{Visualization of results (best viewed in color). We pick 5 representative images from the SUN397 test set and visualize the predicted fine grained categories from our method. We compare them with the predictions from base model (CNN+Softmax). Correct predictions are marked in green, incorrect in red. Failure cases are shown in the rightmost column. "PE" stands for partial evidence.}
\squeezeup
\label{fig:images:sun397}
\end{figure*}

In the second ablation study, we consider different Plugin Network architectures. We evaluate the number of hidden layers of the Plugin Network.
We check from 0 to 4 hidden layers.
The results are in Tab.~\cref{tab:ablation:depth:sun397}.
The best results are obtained by network with 3 hidden layers, which is still a~quite shallow architecture that allows efficient training and does not add significant overhead in the inference.
The results also show that a linear function (model with 0 hidden layers) has not enough capacity to adjust base model outputs.

\begin{table}[t!]
\centering
\begin{tabular}{l|r|r|r|r|r|}
\cline{2-6}
\multirow{2}{*}{} & \multicolumn{5}{c|}{\% of training examples}                                                                                     \\ \cline{2-6} 
                  & \multicolumn{1}{c|}{20} & \multicolumn{1}{c|}{40} & \multicolumn{1}{c|}{60} & \multicolumn{1}{c|}{80} & \multicolumn{1}{c|}{100} \\ \hline
\multicolumn{1}{|c|}{MC Acc}   & 56.58                   & 57.07                   & 57.26                   & 57.34                   & 57.51                    \\ \hline
\multicolumn{1}{|c|}{mAP}      & 60.75                   & 61.19                   & 61.26                   & 61.27                   & 61.37                    \\ \hline
\multicolumn{1}{|c|}{IoU}      & 38.98                   & 39.39                   & 39.42                   & 39.60                   & 39.62                    \\ \hline
\end{tabular}
\caption{Performance of the Plugin Network model trained on a~fraction of training data. The results show that model trained even with 20\% of available data, achieves state-of-the-art performance on all metrics. The results are reported on SUN397 dataset.}
\squeezeup
\label{tab:SUN397:data-amount}
\end{table}

\begin{table}[t!]
\small
\centering
\begin{tabular}{l|r|r|r|}
\cline{2-4}
                       & \multicolumn{1}{c|}{{MC Acc}} & \multicolumn{1}{c|}{{mAP}} & \multicolumn{1}{c|}{{IoU Acc}} \\ \hline
\multicolumn{1}{|c|}{Base model~\cite{Wang18}}               & 52.83\textpm0.24                                & 56.17\textpm0.21                             & 35.90\textpm0.22                                 \\ \hline
\multicolumn{1}{|c|}{SINN \cite{Hu2016, Wang18}}             & 54.30\textpm0.35                                & 58.34\textpm0.32                             & 37.28\textpm0.34                                 \\ \hline
\multicolumn{1}{|c|}{F. Prop (LF) \cite{Wang18}}         & 54.94\textpm0.42                                & 58.52\textpm0.34                             & 37.86\textpm0.39                                 \\ \hline 
\multicolumn{1}{|c|}{F. Prop (RF) \cite{Wang18}}         & 55.01\textpm0.35                                & 58.70\textpm0.26                             & 37.95\textpm0.33                                 \\ \hline \hline
\multicolumn{1}{|c|}{Base (Ours)} & 53.30\textpm0.29                                & 56.36\textpm0.21                             & 34.39\textpm0.31                                 \\ \hline

\multicolumn{1}{|c|}{Plugin Net (Ours)}   & \textbf{57.59\textpm0.24}                                & \textbf{61.55\textpm0.43}                             & \textbf{39.26\textpm0.38}      \\ \hline 

\end{tabular}
\caption{Plugin Network performance on SUN397 dataset. Our method outperforms state-of-the-art on all reported metrics. To allow fair comparison, we also show the performance of base model used in \cite{Wang18} as well as the performance of base model trained by us.}
\squeezeup
\label{tab:results:sun397}
\end{table}

In the third ablation study, we compare different fusion operators from Eq.~\ref{eq:plugin}. The results are presented in Fig.~\ref{img:ablation:FusionOperator:sun397} and show that the best performance and convergence rate is obtained by the additive operator, which is a particular case of affine operator where $\mathbf{r_a} = 1$.
Because we use the ReLU activation function, the translation operation performed by the additive operator is sufficient to alter channel activation values by translating them to either positive or negative half-plane.
Scaling (in the multiplicative operator) can achieve similar effect when $\mathbf{r_a} \approx 0$, but negative $\mathbf{r_a}$
may switch sign of activation values, resulting in amplification of unwanted responses.
Finally we can observe that all proposed fusion operators achieve state-of-the-art results.
As the additive operator achieved consistently the best performance, in all further experiments we are using the additive operator.

In the final model we use AlexNet CNN + Softmax as the~base model. The CNN was pretrained on the Places365 dataset~\cite{Zhou17}.
The base model was chosen to allow a fair comparison with \cite{Wang18, Hu2016}.
We use the Plugin Network with 3 hidden layers, attached to the fc3 layer from the base model.
The model was trained for 15 epochs using the Adam~\cite{kingma2014adam} optimizer with learning rate set to $10^{-3}$,
which was reduced to $10^{-4}$ and $10^{-5}$ after 5 and 10~epochs.

\noindent\textbf{Training analysis:} In Table~\ref{tab:SUN397:data-amount} we report the performance of our method \wrt to the percentage of available training data. We trained Plugin Network with 20\%, 40\%, 60\%, and 80\% percent of the training data. The results in Table~\ref{tab:SUN397:data-amount} show that our model achieves the state-of-the-art performance even when trained on 20\% of the data.

\noindent\textbf{Comparison with the state-of-the-art:} In Table~\cref{tab:results:sun397} we report the performance of our Plugin Network.
The results are averaged over 5 runs to mitigate the randomness in validation set sampling, also standard deviation is computed. 
We report performance of base model, which does not use partial evidence information as a~reference. We also report performance of SINN network \cite{Hu2016} and FeedbackProp \cite{Wang18}.
The results in Table~\cref{tab:results:sun397} show that Plugin Networks outperform state-of-the-art methods in terms of MC Acc, mAP and IoU.
In addition, our method is easier to train than SINN model and allows faster inference than FeedbackProp.

\noindent\textbf{Observations:}
In Fig.~\ref{fig:images:sun397} we show~5 images from the SUN397 test set.
For each example we show: ground-truth label for fine-grained category, both classification results from the base and our model.
We also report coarse category (partial evidence).
The examples show that our method can recover many errors thanks to the presence of partial evidence information.
For instance, in top left example, partial evidence that \say{parking lot} belongs to \say{Outdoor man made} category helped to correct the classification error.
The base model classified example as \say{Anechoic Chamber}, which belongs to \say{Indoor} category.
If we look at the \say{Anechoic Chamber} examples, one can notice that cars in the parking lot can mimic patterns of the \say{Anechoic Chamber} walls.

\begin{figure*}[t]
\centering
\begin{minipage}[b]{0.45\linewidth}
\begin{center}
    \includegraphics[width=0.99\linewidth]{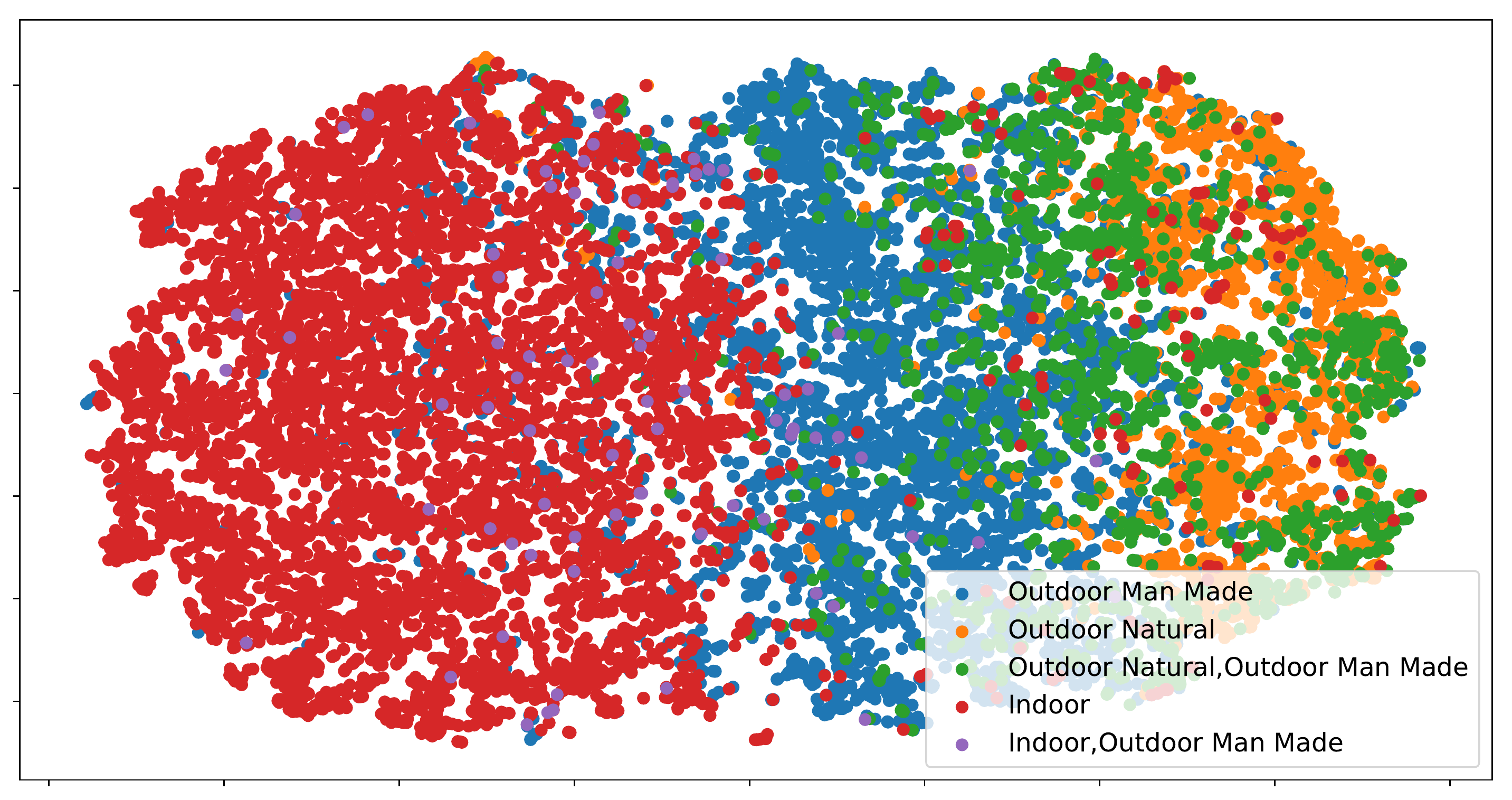} \\
    (a) Base Model
\end{center}
\end{minipage}
\hspace{0.5cm}
\begin{minipage}[b]{0.45\linewidth}
\begin{center}
    \includegraphics[width=0.99\linewidth]{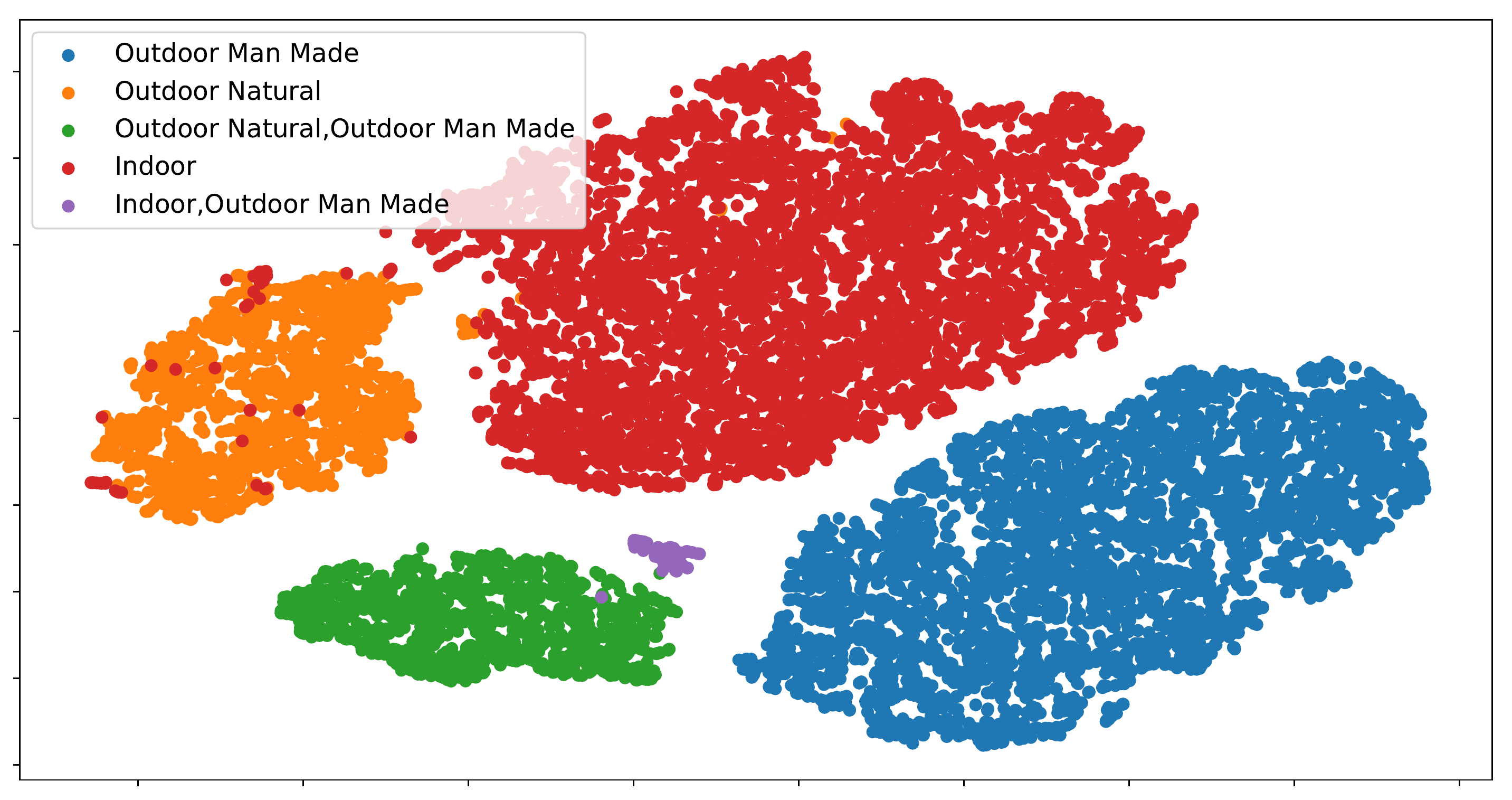} \\
    (b) Plugin Network
\end{center}
\end{minipage}
\caption{Comparison of representation learned by the base model and the Plugin Network. The output of the layer where Plugin Network is attached, is projected to 2 dimensions using t-SNE. Each point in the figure refers to one example from the test set of SUN397. Our model finds better separation \wrt partial evidence categories.}
\label{fig:tsne:SUN397}
\end{figure*}

\noindent\textbf{Representation analysis:}  we analyze how the intermediate output of base model was changed, after Plugin Network was added. First, we projected the activations of base model for each example in a test set to 2D plane using t-SNE method \cite{Maaten08}. Figure \ref{fig:tsne:SUN397} (A) Base Model shows the representation of the base model, while (B) Plugin Network shows the representation of our method. The figure shows that our model learns much better representation, as partial evidence categories are clearly separable. It is also interesting that our model manages to learn such representation, because the loss function that we optimize considers only fine grained labels and ignores error on partial evidence categories. The results show that our model learns dependency between partial evidence categories and fine-grained labels, while not being directly guided by the loss function. 

We also tried to incorporate loss on partial evidence categories, but we did not observe an increase in model performance. Thus we decided to optimize loss function only on unknown labels. Such solution has another advantage -- partial evidence categories do not have to be a subset of labels that base model recognizes.


\subsection{Multi-label Image Annotation}  

In the evaluation of our method for the multi-label image annotation task, we use the COCO 2014 dataset \cite{Lin14}.
It contains 120,000 images, each annotated with 5 caption sentences.
Again, for consistency, we follow the same experimental setup as \cite{Wang18}.
Namely, we use the provided 82,783 training data instances as our training set, and randomly split the remaining provided validation data into 20,000 validation set and 20,504 test set images.

The task is to predict a~predefined set of words explaining an image.
The words are referred as visual concepts in Fang \etal in their visual concept classifier~\cite{Fang15}.
We define them as the 1,000 most frequent words in the captions of the COCO dataset.
We use the same tokenization, lemmatization, and stop-word removals as Wang \etal~\cite{Wang18}.
As a~result, each image is annotated by a~vector of 1,000 elements corresponding to an occurrence of words in the captions.

For the task of reasoning under partial evidence, we randomly divide the target vector into a~fixed 500 {known} and 500 {unknown} classes.
The model performance is measured on the {unknown set} only, while the {known set} is used as the partial evidence.
The base network is first trained as in Fang \etal~\cite{Fang15} on the multi-labeled task using the entire 1,000 classes.
It is done by minimization of the binary cross entropy between the predicted and target vector of concepts.
For this experiment, for the base network we choose to use the ResNet-18 architecture~\cite{he2016deep} to stay consistent with~\cite{Wang18}.
Additionally, we also make an ablation study for deeper base network architectures: ResNet-50 and ResNet-101.

\begin{figure}[t!]
\begin{center}
    \includegraphics[width=0.95\linewidth]{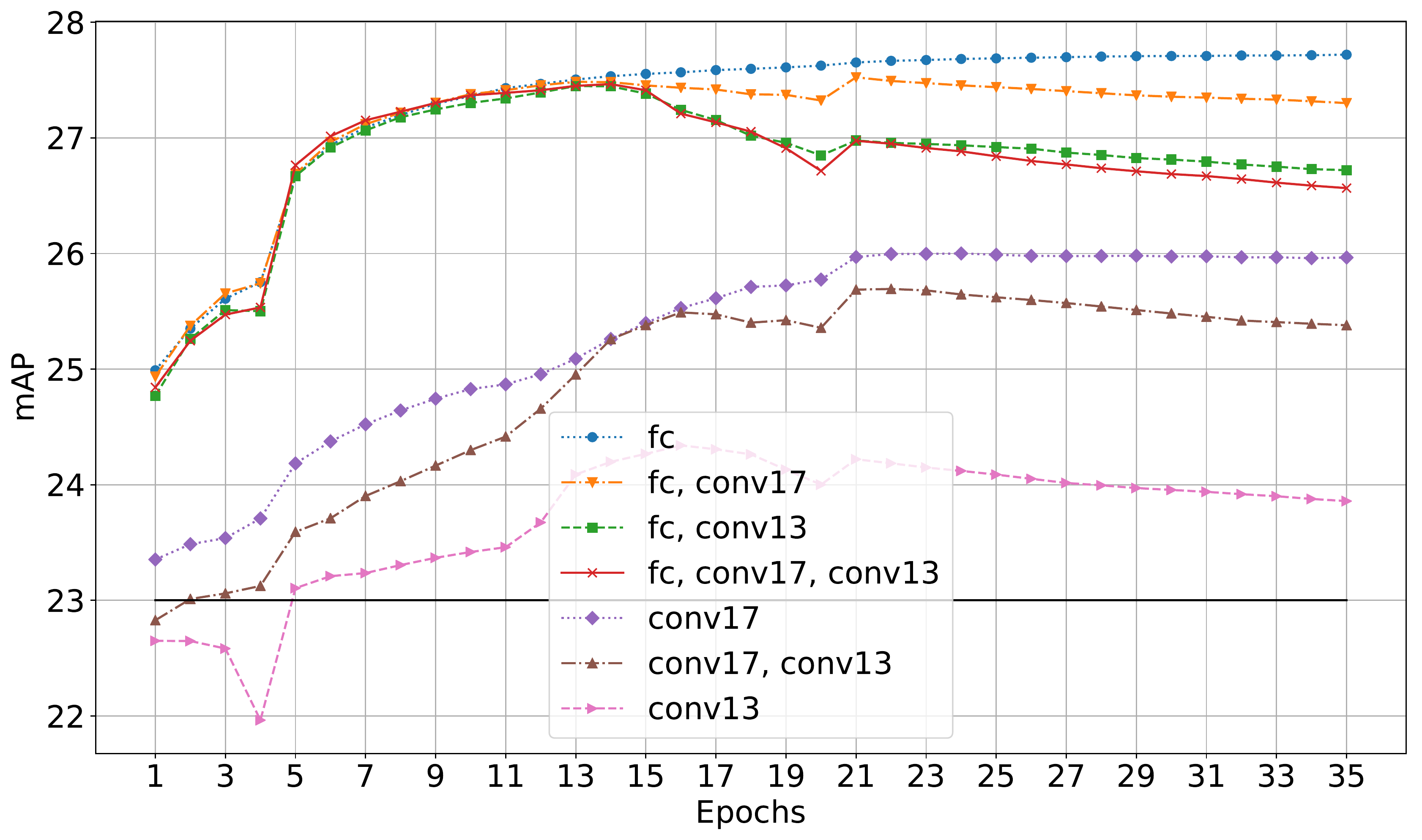}
\end{center}
\vspace{-1.5em}
\caption{Ablation study of Plugin Network attachment layer. Base network is ResNet-18.
Plugins are attached to 17th and 13th conv layer and the last FC layer.
The plugins architecture, consists of 2 layered FC network with 500, and 2048 neurons.
The black solid line indicates mAP achieved by ResNet-18 w/o Plugin Network.
Attachment to the last fc layer only results in the highest improvement.
}
\squeezeup
\label{fig:connection-check}
\end{figure}

\begin{figure}
\begin{center}
    \includegraphics[width=0.95\linewidth]{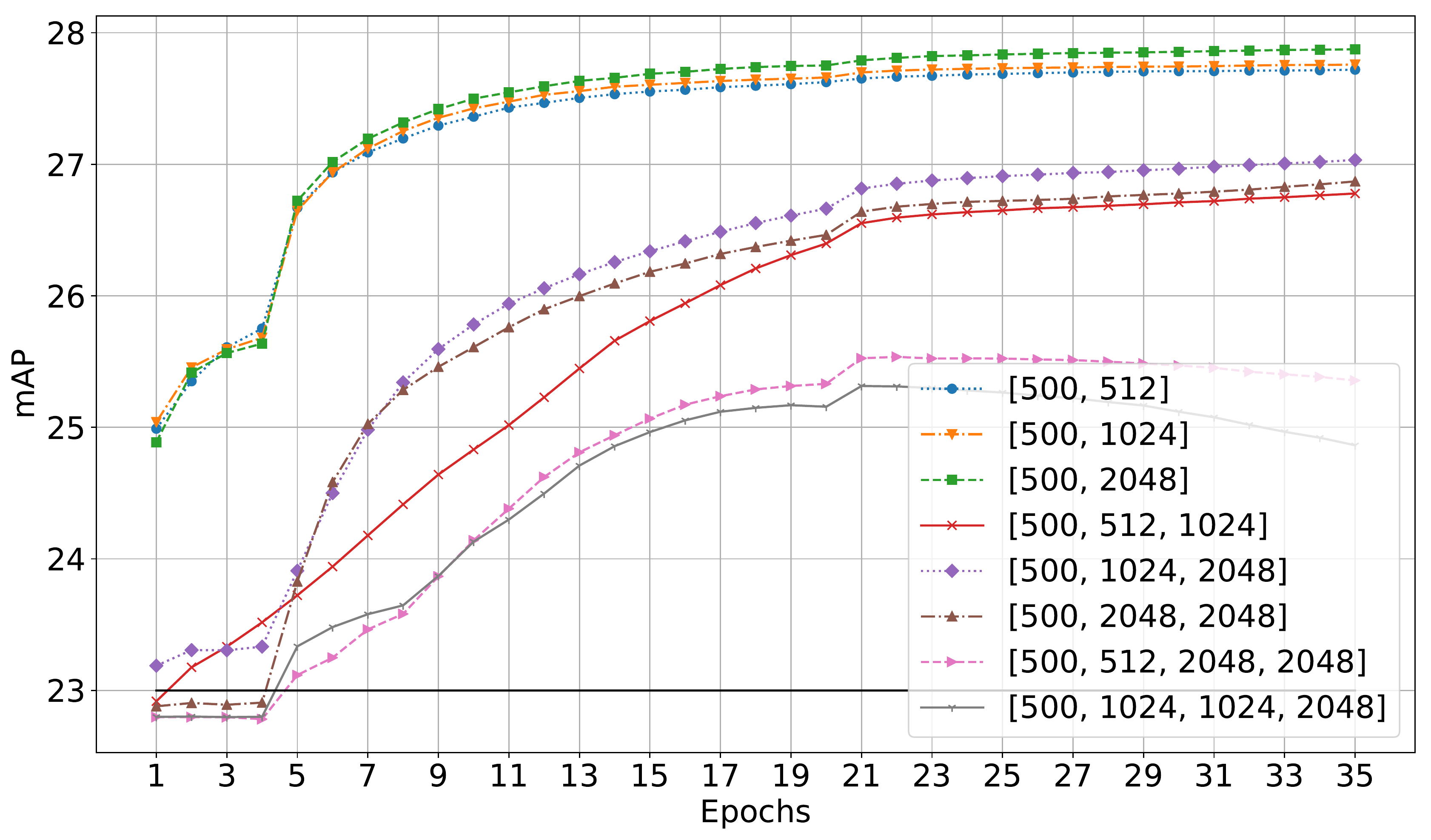}
\end{center}
\vspace{-1.5em}
\caption{
Ablation study of Plugin Network architecture. Base network is ResNet-18. Plugin network is attached to the last layer. Numbers in brackets indicate the number of neurons at consecutive layers.
The black solid line mAP achieved by ResNet-18 w/o Plugin Network.
If the plugin has too many layers, it starts to overfit.
We report the highest performance by using a~two-layered network.
}
\squeezeup
\label{fig:architecutre-check}
\end{figure}

\noindent\textbf{Hyperparameters selection:}
The architecture of the plugin has been chosen based on the validation scores from Figs.~\cref{fig:connection-check} and \cref{fig:architecutre-check}.
As indicated before, we first train the base network on the given task.
Then, we freeze the base network's weights and add a~number of plugins.
We train each plugin for 36 epochs with a~starting learning rate of $1\mathrm{e}{-3}$, which decrease with the number of epochs to  $1\mathrm{e}{-4}$.
We use the Adam~\cite{kingma2014adam} optimizer with the Xavier initialization~\cite{glorot2010understanding}.


During the hyperparameters selection, we verified all combinations of attaching a~plugin to the {conv13}, {conv17} and FC layers of the ResNet-18.
We report, that a~single attachment to the last FC layers results in the highest mAP.
Using any earlier layer still improves the baseline, but such plugin overfits much easier.
Using a~combination of the FC layer and any other convolutional layer leads to a~lower performance comparing to a~single attachment to the FC layer. due to the overfitting;[MK] 
see Fig.~\cref{fig:connection-check}.
This results are aligned with the conclusions 
from Section \ref{sec:sun397}.

In next experiment we search for the best architecture of a single plugin.
We consider different number of layers and neurons in the Plugin Networks.
See Fig.~\cref{fig:architecutre-check} for details.
The highest score is achieved by using the two layered architectures with 500 and 2048 neurons respectively.
The two and three layered networks get to the plateau after around 20 epochs, while the four-layered networks start to overfit.
\begin{table}[t!]
\centering
\begin{tabular}{l|c|c|c|c|}
\cline{2-5}
    &\multicolumn{1}{c|}{\parbox[t]{1cm}{\centering no\\[-2pt]plugin}} &\multicolumn{1}{l|}{\parbox[t]{1.5cm}{\centering FC,\\[-2pt]RL4, RL3}} &\multicolumn{1}{c|}{\parbox[t]{0.5cm}{FC,\\[-2pt]RL4}} &\multicolumn{1}{c|}{\textbf{{FC}}} \\ \hline
  \multicolumn{1}{|c|}{ResNet18} & 23.00             &  27.56    &  27.61  & \textbf{27.97}     \\ \hline
  \multicolumn{1}{|c|}{ResNet50} & 25.84             &  29.85    &  29.65  & \textbf{29.93}    \\ \hline
  \multicolumn{1}{|c|}{ResNet101} & 26.56            &  29.49    &  29.79  & \textbf{30.13}     \\ \hline 
\end{tabular}
\caption{
Scores obtained when attaching plugins at different places to different versions of ResNet.
We consider attachments to the last FC layer, end of the third residual layer (RL3) and fourth residual layer (RL4).
Plugins always consist of 500 and 2048 neurons.
For this task, we always achieve the highest score when applying a~single~plugin to the last layer of the base network.}
\squeezeup
\label{tab:results:resnet100}
\end{table}
\begin{table}[t!]
\centering
\begin{tabular}{l|c|c|c|}
\cline{2-3}
    &\multicolumn{1}{l|}{{mAP}} &\multicolumn{1}{l|}{{Inference time [s]}} \\ \hline
   \multicolumn{1}{|c|}{Base model \cite{Fang15}}          & 23.00             & 25.64             \\ \hline
\multicolumn{1}{|c|}{F. Prop (LF) \cite{Wang18}}     & 25.26             & 93.36             \\ \hline
\multicolumn{1}{|c|}{F. Prop (RF) \cite{Wang18}}     & 25.70             & 103.27             \\ \hline \hline 
\multicolumn{1}{|c|}{Plugins (Ours)}       & \textbf{27.97}             & \textbf{25.72}             \\ \hline
\end{tabular}
\caption{Results on the COCO'14. The baseline is ResNet-18 trained for the multi-label experiment.
Our method not only achieves the state-of-the-art in means of the mAP, but also is the fastest during inference, being barely slower than the base network.}
\squeezeup
\label{tab:results:coco}
\end{table}

\noindent\textbf{Comparison with the state-of-the-art:}
We compare ourselves to the Layer-wise Feedback-prop (LF) and Residual Feedback-prop (RF) Inference proposed by \cite{Wang18}.
Results presented in this work are based on the open sourced online implementation\footnote{\url{github.com/uvavision/feedbackprop}}  provided by the authors of RF and LF. 
Due to the random choice of the known and unknown labels, the baseline may differ, but the overall gain stays similar.
We show, that in terms of the mAP, we achieve the state-of-the-art with a~significant margin.
Furthermore, as expected, the inference phase is much faster compared to the Feedback-prop methods.
Please refer to Tab.~\cref{tab:results:coco} for results. 

Finally, we verify the usage of the Plugin Networks for deeper architectures, such as ResNet50 and ResNet101; see~Tab.~\cref{tab:results:resnet100}.
We notice an improvement for each base network.
As for the ResNet-18, we achieve the highest score when only one plugin is used at the last FC layer.

In this section we evaluate the Plugin Networks on multi-cue object class segmentation task.
We take fully convolutional network (FCN)~\cite{Long-FCN-2015} as a~starting point.
Next, we experiment by adding several plugins into its architecture.
For the baseline we take the pre-trained FCN-8s model\footnote{\url{github.com/wkentaro/pytorch-fcn}} trained on the SBD dataset \cite{Hariharan-SBD-2011}.
We use the same dataset when training the Plugin Networks.
We validate our models on the Pascal VOC 2011 segmentation challenge~dataset~\cite{pascal-voc-2011}.
We follow~\cite{Long-FCN-2015} and take \textit{Pascal VOC 2011 segval\footnote{\url{github.com/shelhamer/fcn.berkeleyvision.org}}}
validation split in order to avoid overlapping images between these two datasets.
Thus, our training and validation datasets consist of 8498 and 736 images, respectively.
Objects in the image are assigned to one of 21 classes.

The base model (FCN-8s without plugins) results in IoU score of 65.5\%.
For this scenario, we assume that we have the knowledge of classes present in the image at the inference time, which constitutes partial evidence.
Therefore, our partial evidence is a~vector of 21 elements.
The goal of the Plugin Network is to improve the output segmentation masks of the base network.
We experiment with attaching several plugins to the FCN-8s model. 
We report the results in Tab.~\cref{tab:segmentation}.
In contrary to findings from previous experiments, the Plugin Networks provide the highest increase in the IoU, when multiple of them are used.
We achieve the highest gain of $72.2\%$ of IoU when attaching 5 plugins into all of the convolutional layers and all of the transposed convolutional layers.
We also outperform the previously proposed method~\cite{Rosenfeld18}, which achieved $69.2\%$.

\subsection{Scene semantic segmentation}
\label{sec:semantic:segmentation}
\begin{table}[]
    \centering
    \begin{tabular}{|c|c|c|} 
    \hline
    Model & \# of plugins & mean-IoU \\ \hline \hline
    Baseline & 0 & 65.5 \\ \hline  \hline
    conv1-3 & 3 & 65.7 \\ \hline
    conv1-5 & 5 & 70.5 \\ \hline
    deconv1-3 & 3 & 71.1 \\ \hline
    deconv1-5 & 5 & 71.2 \\ \hline
    conv1-5, deconv1-5 & 10 & \textbf{72.2} \\ \hline
    \end{tabular}
    \caption{
    Scores obtained when attaching plugins at different layers of the FCN architecture for the task of semantic segmentation. conv stands for a convolutional and deconv for transposed convolutional layers.
    }
    \squeezeup
    \label{tab:segmentation}
\end{table}
\begin{figure}[t!]
\begin{center}
    \includegraphics[width=0.99\linewidth]{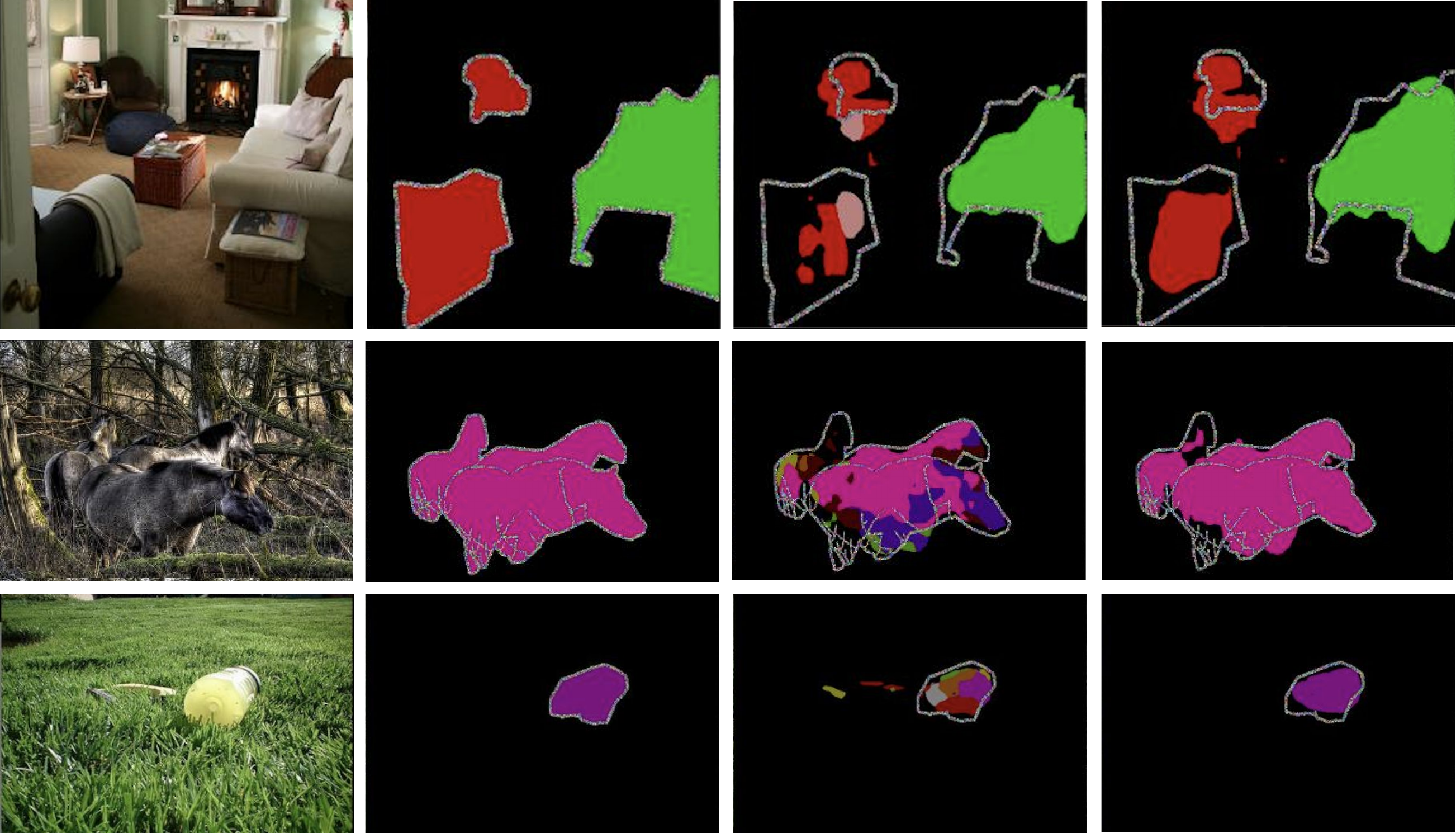}
\end{center}
\vspace{-1.5em}
\caption{
Examples of semantic segmentation masks predicted by the base model with and without Plugin Networks. The images are taken from the Pascal VOC 2011 validation set.
In rows we show different examples, and in columns, from left to right: input image, ground-truth (all classes), base model, FCN-8s with 10 plugins. 
Our method show clear improvement of the output segmentation masks.
}
\squeezeup
\label{fig:seg}
\end{figure}

Using partial evidence through the Plugin Networks, we are able to soften the wrong feature maps and strengthen the expected ones.
It results in major improvement of the base network.
When multiple objects are present in the scene, the base model may have a~problem to consistently assign a~proper label to a~particular object.
As expected, when the baseline network makes a~mistake by assigning a~wrong label of a~class that is not present in the image, plugins correct these with a~correct class.
The examples are shown in Fig.~\cref{fig:seg}.
For instance in the last row, the pixels belonging to a bottle are inconsistently assigned to different classes by the base model.
Using the Plugin Networks fixes the problem. 

\section{Conclusions}
\label{sec:conclusions}
In this work, we introduced the Plugin Networks -- a~simple, yet effective method to exploit the availability of a partial evidence in the context of visual recognition tasks. Plugin Networks are integrated directly with the intermediate layers of pre-trained convolutional neural networks and thanks to their lightweight design can be trained efficiently with low computational cost and limited amount of data. Results presented on three challenging tasks and various datasets show superior performance of the proposed method with respect to the state-of-the-art approaches. 

We believe that this work can open novel research directions related to solving visual recognition tasks with partial evidence, as our Plugin Networks are agnostic to the input signal and can accommodate arbitrary modality of the input data, including audio or textual cues. Therefore, their multimodal nature can allow richer contextual cues to be taken into account in the inference procedure, leading to more effective and efficient visual recognition models.

\section*{Acknowledgements}
\label{sec:acknowledgements}
\begin{small}
This research was partially supported by the Polish National Science Centre grant no. UMO-2016/21/D/ST6/01946.
\end{small}

{\small
\bibliographystyle{ieee}
\bibliography{egbib}
}

\end{document}